\documentclass[conference]{IEEEtran}
\IEEEoverridecommandlockouts
\usepackage{cite}
\usepackage{amsmath,amssymb,amsfonts}
\usepackage{algorithm}
\usepackage[noend]{algpseudocode}
\usepackage{graphicx}
\usepackage{textcomp}
\usepackage{xcolor,subcaption}
\begin{document}
\makeatletter
\newcommand{\linebreakand}{%
  \end{@IEEEauthorhalign}
  \hfill\mbox{}\par
  \mbox{}\hfill\begin{@IEEEauthorhalign}
}
\makeatother

\title{HTPS: Heterogeneous Transferring Prediction System for Healthcare Datasets}

\author{\IEEEauthorblockN{1\textsuperscript{st} Jia-Hao Syu}
\IEEEauthorblockA{~~~~~~~~~~\textit{Department of Computer Science}~~~~~~~~~~ \\
\textit{and Information Engineering}\\
\textit{National Taiwan University}\\
Taipei, Taiwan \\
f08922011@ntu.edu.tw}
\and
\IEEEauthorblockN{2\textsuperscript{nd} Jerry Chun-Wei Lin}
\IEEEauthorblockA{\textit{Department of Computer Science,} \\
\textit{Electrical Engineering and Mathematical Sciences}\\
\textit{Western Norway University of Applied Sciences}\\
Bergen, Norway \\
jerrylin@ieee.org}
\linebreakand 
\IEEEauthorblockN{3\textsuperscript{rd} Marcin Fojcik}
\IEEEauthorblockA{\textit{Department of Computer Science,} \\
\textit{Electrical Engineering and Mathematical Sciences}\\
\textit{Western Norway University of Applied Sciences}\\
Bergen, Norway \\
Marcin.Fojcik@hvl.no}
\and
\IEEEauthorblockN{4\textsuperscript{th} Rafał Cupek}
\IEEEauthorblockA{~~~~~~~~~~\textit{Department of Distributed Systems}~~~~~~~~~~ \\
\textit{and Informatic Devices}\\
\textit{Silesian University of Technology}\\
Gliwice, Poland \\
Rafal.Cupek@polsl.pl}
}

\maketitle

\begin{abstract}
Medical internet of things leads to revolutionary improvements in medical services, also known as smart healthcare.
With the big healthcare data, data mining and machine learning can assist wellness management and intelligent diagnosis, and achieve the P4-medicine.
However, healthcare data has high sparsity and heterogeneity.
In this paper, we propose a Heterogeneous Transferring Prediction System (HTPS).
Feature engineering mechanism transforms the dataset into sparse and dense feature matrices, and autoencoders in the embedding networks not only embed features but also transfer knowledge from heterogeneous datasets.
Experimental results show that the proposed HTPS outperforms the benchmark systems on various prediction tasks and datasets, and ablation studies present the effectiveness of each designed mechanism.
Experimental results demonstrate the negative impact of heterogeneous data on benchmark systems and the high transferability of the proposed HTPS.
\end{abstract}


\section{Introduction}
\IEEEPARstart{W}{ith} the continuous development of Medical Internet of Things (MIoT), MIoT has revolutionary improvements in medical services, known as smart healthcare~\cite{MIoT}.
Smart healthcare involves the development of smart hospitals, IoT sensing, and diagnosis assisting.
Digital information assists hospitals in management decision-making and nursing process improvement, and virtualized services are exhibited in the fields of remote clinics and medical consultation, which increase the supply and alleviate the demand for medical services~\cite{RC1}.
MIoT sensing could monitor respiration, heart rate, blood pressure, and even mental stress and mood through home medical instruments and wearable devices (such as smartphones and smartwatches).
With the big healthcare data, artificial intelligence (AI) technologies of data mining and machine learning can assist wellness management and intelligent diagnosis, and achieve the P4-medicine (preventive, participatory, predictive, and personalized)~\cite{P4}.

However, healthcare data has high sparsity and heterogeneity because of different users, devices, and frequencies.
For example, ICU patients, hospitalized patients, and healthy users have different health conditions, and different instruments may be used to measure indicators in time frequencies.
In addition, healthcare data collection is limited by various issues; therefore, a single source could only collect limited data.
Therefore, how to make full use of heterogeneous healthcare data is still an unsolved problem.

To ease the issues, we propose a Heterogeneous Transferring Prediction System (HTPS).
The feature engineering mechanism of HTPS transforms the dataset into sparse and dense feature matrices, which are inputs of the designed sparse and dense embedding networks, respectively.
The autoencoders in the dense embedding networks not only embed the features but also transfer knowledge from heterogeneous models with different datasets and prediction tasks.
The contributions of this paper can be summarized as follows:
\begin{enumerate}
\item Feature engineering mechanism extracts feature-based information (dense feature matrix) and time-series-based information (sparse feature matrix).
\item Embedding networks not only embed features but also enhance regularization as multi-task learning.
\item Heterogeneous transfer learning mechanism transfers knowledge from source model with different datasets, features, and prediction tasks.
\end{enumerate}
We evaluate the proposed HTPS on various prediction tasks and datasets, which outperforms the benchmark systems.
Ablation studies are executed to present the effectiveness of each designed mechanism in HTPS.
Experimental results demonstrate the negative impact of heterogeneous data on benchmark systems and the high transferability of the proposed HTPS.

\section{Literature Review}   \label{sec:LR}
In this section, we review the literature on smart healthcare in Section~\ref{sec:LR-SH}.
Section~\ref{sec:LR-HD} introduces the heterogeneous data.

\subsection{Smart Healthcare}   \label{sec:LR-SH}
Smart healthcare aims to provide people with convenient and personalized services through the technologies of artificial intelligence~\cite{SH}.
The research of smart healthcare includes smart hospitals, MIoT sensing, and diagnosis assisting.

With the approach of a super-aged society, the demand for healthcare services has grown gradually, which triggers discussions on smart hospitals~\cite{SHospital}.
Digital information assists hospitals in management decision-making and nursing process improvement.
Virtualized services are exhibited in the fields of remote clinics and medical consultation.
During the COVID-19 epidemic, virtualized medical services have greatly alleviated the soaring medical demand~\cite{RC1,RC2}.
In addition, remote medical services require low-latency and secure communication, leading to discussions of communication standards and algorithms for healthcare applications~\cite{HComm}.

MIoT sensing could monitor respiration, heart rate, blood pressure, and even mental stress and mood through home medical instruments and wearable devices, and enables non-critical patients to recover outside the hospital.
Through data mining analysis and machine learning prediction, smart and wearable devices can suggest human living habits, making IoT go beyond monitoring, but wellness management and intelligent diagnosis.
For example, predictive data mining methods can diagnose heart disease and diabetes~\cite{DMH}, and deep neural networks can assist in identifying and classifying various cancers~\cite{Cancer}.
However, as more healthcare data are collected and decision-making recommendations are provided, privacy and security issues should be given great attention, and could be alleviated by distributed learning algorithms.

\subsection{Heterogeneous Data}   \label{sec:LR-HD}
Besides learning from horizontally and vertically distributed data, data is heterogeneous if it is collected from different populations with different distributions.
Since heterogeneous data have different sizes, distributions, and feature types, it can negatively impact model performance for distributed learning, such as prediction ability~\cite{HData1} and convergence ~\cite{HData2}.
To make full use of knowledge from heterogeneous data, several algorithms for heterogeneous data are proposed based on meta-learning~\cite{MetaL} and personalized federated learning~\cite{PFL}.

\section{Proposed Heterogeneous Transferring Prediction System (HTPS)}   \label{sec:Meth}
In this section, the proposed HTPS is introduced, including three mechanisms of feature engineering, neural network prediction, and heterogeneous transferring, detailed in Sections~\ref{sec:Meth-Fea} to~\ref{sec:Meth-HT}.

\subsection{Feature Engineering}   \label{sec:Meth-Fea}
Traditional machine learning algorithms take features at time-series $t_{1}, t_{2}, \ldots, t_{W}$ as input, and any pair of two adjacent timestamps has the same distance.
However, healthcare data has high sparsity because not all features are measured and collected at the same time-frequency.
Unstable and sparse features have a negative impact on neural network predictions.
To deal with the sparsity, the feature engineering mechanism in the HTPS transforms the healthcare data into a sparse and a dense feature matrix.

In the feature engineering mechanism, the timestamp does not depend on a constant time slot but on the feature collected order.
Suppose there are $M$ records for a user, sorted by collected time.
Each record has only a value of a specific feature, and there are $N+1$ different features in total (1 predicted target and N predicted features).
All the sparse and dense feature matrices are in the shape of $W\times N$, where $W$ is the window size.

The sparse feature matrices for records, $R_{1}, \ldots, R_{M}$, are generated as follows.
For each record $R_{t}$, the algorithm extracts the record as feature type $FT$ and feature value $FV$, where $FT \in \{0, \ldots, N\}$ and the prediction target has a $FT$ of 0.
If $FT \neq 0$, the record $R_{t}$ is transformed to a sparse record $SR$ by Sparse($\cdot$), which generates an $N$-dimensional zero vector with the $FT$-th dimension being $FV$.
The last $W$ sparse records are stored in $SRL$.
If $FT = 0$ and $SRL$ contains $W$ sparse records, generate a new sparse feature matrix, and add it to the list of sparse feature matrices $SFM$.
Furthermore, the corresponding label (prediction target) $FV$ is added to the label list $L$.
After traversing all records $R_{t}$, the algorithm finally outputs the list of sparse feature matrices $SFM$ and the corresponding label list $L$.
To sum up, the sparse feature matrices store the last $W$ measurements, and only contain $W$ values in a $W\times N$ matrix.

The dense feature matrix for records, $R_{1}, \ldots, R_{M}$, are generated as follows.
For each record $R_{t}$, the algorithm extracts the record as feature type $FT$ and feature value $FV$.
If $FT \neq 0$, the value $FV$ is stored in the $FT$-th dimension of the dense record list $DRL$.
Similarly, $DRL$ only stores the last $W$ values of each feature.
If $FT = 0$ and all $N$ dimensions of $DRL$ are $W$, generate a new dense feature matrix, and add it to the list of dense feature matrices $SDM$.
The corresponding label (prediction target) $FV$ is also added to the label list $L$.
After traversing all records $R_{t}$, the algorithm finally outputs the list of dense feature matrices $SDM$ and the corresponding label list $L$.
In summary, the dense feature matrices store the last $W$ measurement information of all $N$ features.

\begin{algorithm}[t]
\footnotesize 
  \caption{Sparse Feature Matrix}
  \label{alg:SFM}
  \begin{algorithmic}[1]
    \State  $SFM$, $L$, $SRL$ = [], [], []
    \For{$t = 1, \ldots, M$}
    	\State $FT$, $FV$ = extract($R_{t}$)
    	\If{$FT$ = 0 \& len($SRL$) $= W$}
    		\State $SFM$ += $SRL$
    		\State $L$   += $FV$
    	\EndIf
    	\If{$FT \neq$ 0}
    		\State $SRL$ += Sparse($FT$, $FV$)
    		\State Save the last $W$ records
    	\EndIf
    \EndFor
    \State Return $SFM$, $L$
  \end{algorithmic}
\end{algorithm}
\begin{algorithm}[t]
\footnotesize 
  \caption{Dense Feature Matrix}
  \label{alg:DFM}
  \begin{algorithmic}[1]
    \State  $DFM$, $L$, $DRL$ = [], [], [[] $\times$ $N$]
    \For{$t = 1, \ldots, M$}
    	\State $FT$, $FV$ = extract($R_{t}$)
    	\If{$FT$ = 0 \& $\forall i$, len($DRL[i]$) $= W$}
    		\State $DFM$ += $DRL$
    		\State $L$   += $FV$
    	\EndIf
    	\If{$FT \neq$ 0}
    		\State $DRL$[FT] += $FL$
    		\State Save the last $W$ records
    	\EndIf
    \EndFor
    \State Return $DFM$, $L$
  \end{algorithmic}
\end{algorithm}

\subsection{Neural Network Prediction}   \label{sec:Meth-Net}
\begin{figure*}[t] 
  \centering
  \includegraphics[width=6.7in]{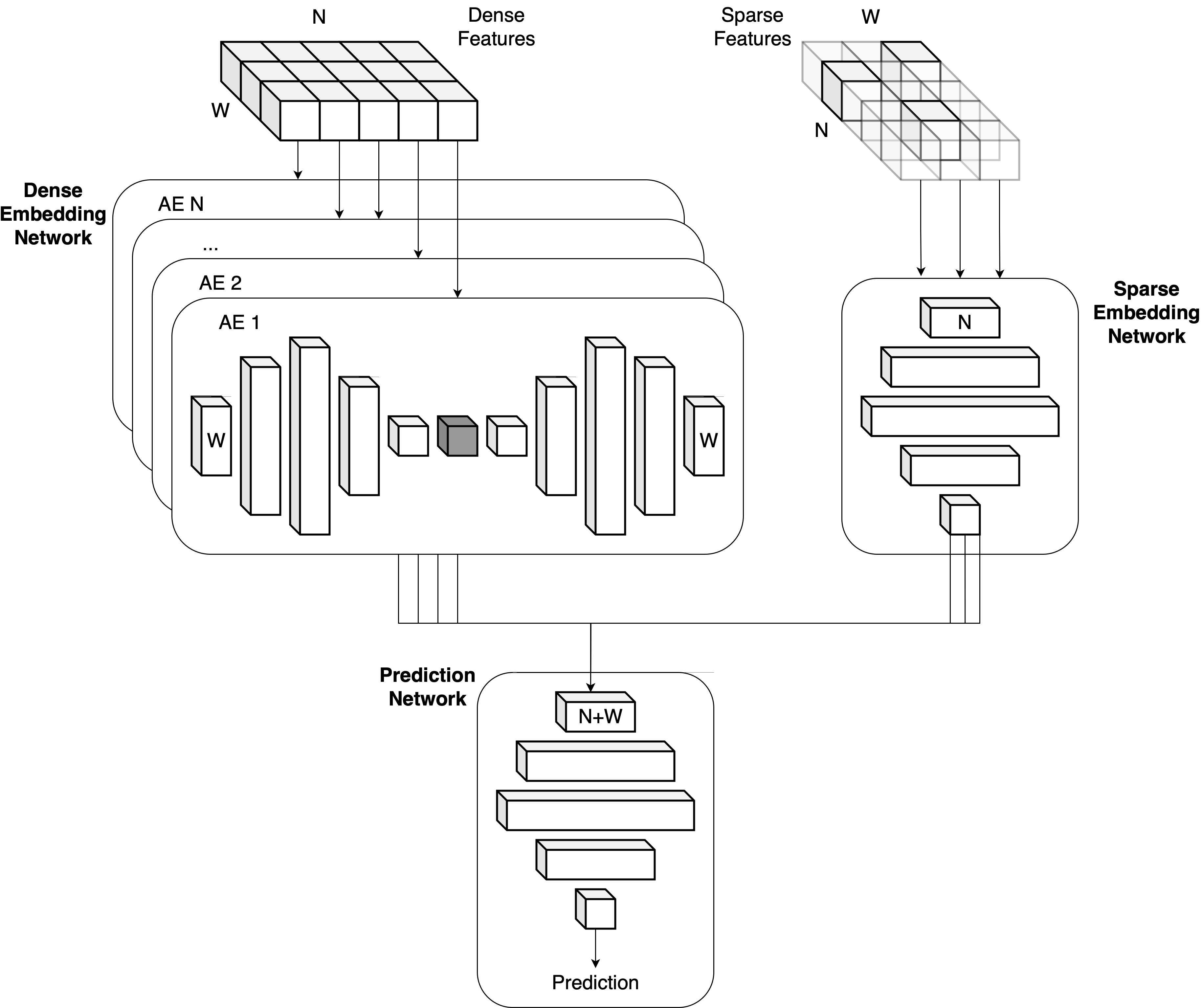}
  \caption{Schematic diagram of the HTPS network design}
  \label{fig:ND}
\end{figure*}
The flow chart of the designed network is presented in Fig.~\ref{fig:ND}, including 3 sub-networks of dense embedding network, sparse embedding network, and prediction network.
The dense embedding network embeds information of $N$ features separately (for heterogeneous transferring), while the sparse embedding network embeds time-series information in the last $W$ records.
The prediction network collects embedding vectors to make predictions.

In the dense embedding network, there are $N$ autoencoder networks for $N$ features, respectively.
The autoencoders have symmetric pairs of encoder and decoder, which consists of fully connected layers), as shown in Fig.~\ref{fig:AE}.
An encoder takes a feature in the dense feature matrix as ($W$-dimensional) input, and embeds it into a 1-dimensional vector.
The decoder takes the 1-dimensional vector as input to reconstruct the $W$-dimensional feature.
We adopt mean absolute error (MAE, L1 loss) to evaluate the performance of autoencoders.
To sum up, the dense embedding network takes $W \times N$-dimensional dense feature matrix as input, and splits the matrix into $N$ $W$-dimensional vectors for embedding, and transmits the $N$ embedded vectors to the prediction network.

The sparse embedding network is designed as fully connected layers with an input size of $N$ and an output size of $1$.
The sparse embedding network splits the sparse feature matrix into $W$ $N$-dimensional vectors, sequentially fed into the layers to obtain $W$ 1-dimensional embedded vectors.
To sum up, the sparse embedding network embeds time-series information in the last $W$ records.

The prediction network collects $N$ 1-dimensional embedded vectors from the dense embedding network and $W$ 1-dimensional embedded vectors from the sparse embedding network as input, and predicts the target value through fully connected layers.
In the HTPS, there are $N+1$ losses, which are $N$ MAE of autoencoders in the dense embedding network and a mean square loss (MSE) of the final predicted value.
The numbers of neurons of layers and networks are listed in Table~\ref{tab:NN}.
Note that the number of neurons may be slightly adjusted to match the number of parameters of the benchmark system.
\begin{figure}[t] 
  \centering
  \includegraphics[width=3.5in]{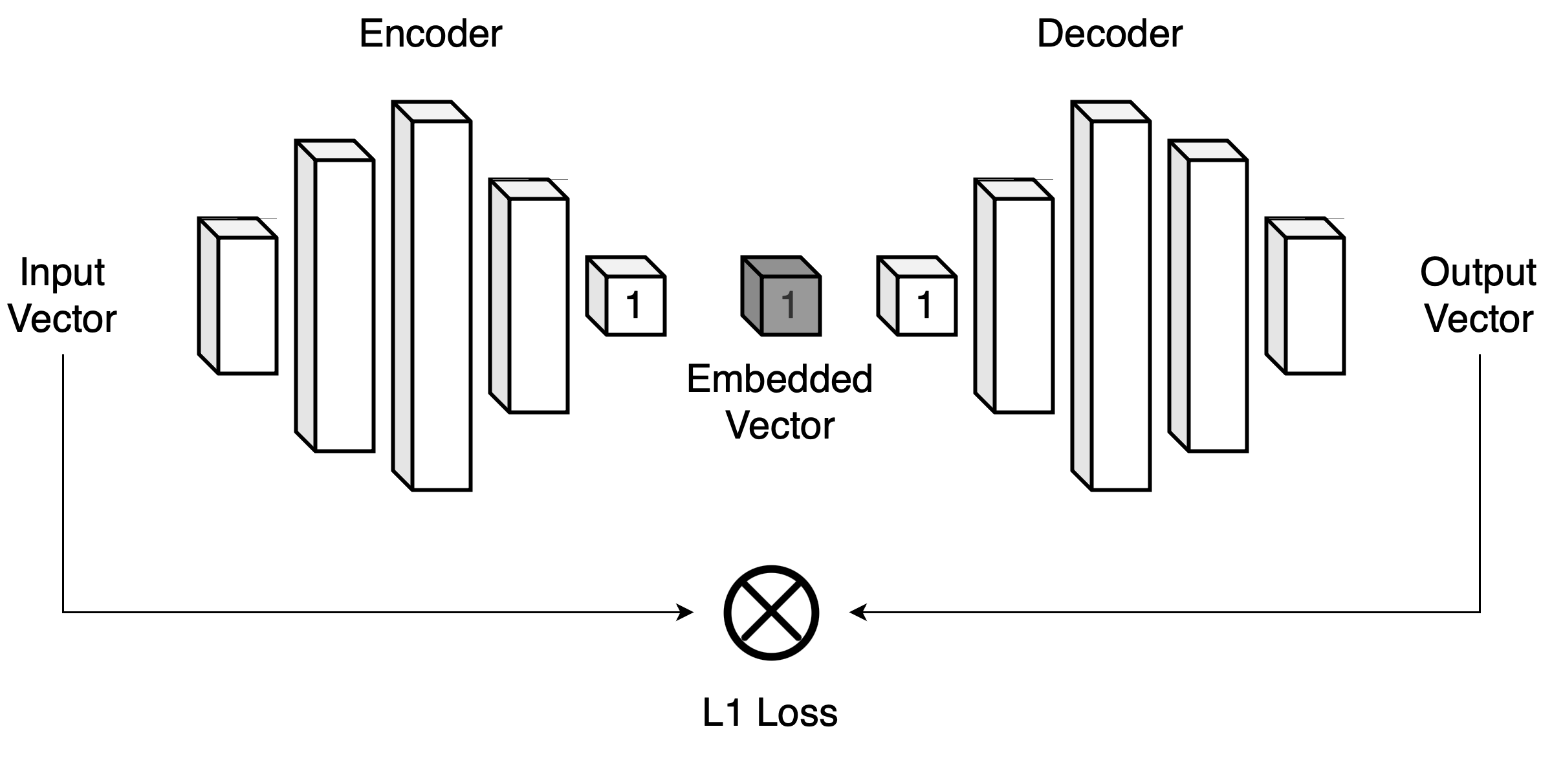}
  \caption{Autoencoder}
  \label{fig:AE}
\end{figure}
\begin{table}[t]
\caption{Details of networks design}
\centering
\begin{tabular}{|lc|lc|}
\hline
\multicolumn{2}{|c|}{Dense Network - Encoder} & \multicolumn{2}{c|}{Dense Network - Decoder}\\
\hline
Layer  & Neurons & Layer  & Neurons \\
\hline
Linear & $W$ & Linear &   6\\
LeakyReLU &  & LeakyReLU & \\
Linear &  32 & Linear & 256\\
LeakyReLU &  & LeakyReLU & \\
Linear & 256 & Linear &  32\\
LeakyReLU &  & LeakyReLU & \\
Linear &   6 & Linear & $W$\\
LeakyReLU &  & & \\
Linear &   1 & &\\
\hline
\multicolumn{4}{c}{}\\
\hline
\multicolumn{2}{|c|}{Sparse Embedding Network} & \multicolumn{2}{c|}{Prediction Network}\\
\hline
Layer  & Neurons & Layer  & Neurons \\
\hline
Linear & $N$ & Linear & $N$ + $W$\\
LeakyReLU &  & LeakyReLU & \\
Linear &  32 & Linear &  32\\
LeakyReLU &  & LeakyReLU & \\
Linear & 256 & Linear & 256\\
LeakyReLU &  & LeakyReLU & \\
Linear &   6 & Linear &   6\\
LeakyReLU &  & LeakyReLU & \\
Linear &   1 & Linear &   1 \\
\hline
\end{tabular}
\label{tab:NN}
\end{table}

\subsection{Heterogeneous Transferring}   \label{sec:Meth-HT}
Datasets from different sources may have different features.
How to transfer knowledge between heterogeneous datasets is a critical issue in healthcare research.
To make full use of heterogeneous datasets, a heterogeneous transferring mechanism is designed in this section.

Let the source domain be the knowledge (dataset and model) transferred from, and the target domain be the knowledge transferred to.
Also, let the source model $\Bbb M^{S}$ be the model (in Section~\ref{sec:Meth-Net}) trained on the source dataset.
The designed heterogeneous transfer learning mechanism aims to transfer the knowledge in $\Bbb M^{S}$ to the target model $\Bbb M^{T}$ as initialization.

There may be a different number of features in the source dataset and target dataset, resulting in a different number of autoencoders in the dense embedding network.
Therefore, model weights cannot be transferred (copied) directly.
We design a feature matching mechanism as follows.
Suppose there are $N^{S}$ and $N^{T}$ features (autoencoders) in the source and target dataset.
For each feature in the target dataset, the corresponding vectors in the dense feature matrices are fed to the $N^{S}$ autoencoders in $\Bbb M^{S}$ to obtain $ N^{S}$ MAE losses.
The feature will match the autoencoder in $\Bbb M^{S}$ with the lowest MAE error, which's weights are transferred (copied) to the corresponding autoencoder in $\Bbb M^{T}$  as initialization.
The feature matching mechanism is applied to all $N^{T}$ features in the target dataset and only on the training set.
In summary, the heterogeneous transferring finds the autoencoder in $\Bbb M^{S}$ with the lowest reconstruction error for each feature in the target dataset and transfer weights of autoencoders in $\Bbb M^{S}$ to $\Bbb M^{T}$.

\section{Experimental Results and Discussion}   \label{sec:Exp}
In this section, the dataset usage and benchmark systems are introduced in Section~\ref{sec:Exp-Data}.
Section~\ref{sec:Exp-PE} then presents the prediction evaluation, and Section~\ref{sec:Exp-AS} present the ablation studies.

\subsection{Dataset Usage and Benchmark Systems}   \label{sec:Exp-Data}
In this paper, MIMIC-III clinical database~\cite{MIMIC-2} is adopted to evaluate the effectiveness in the following experiments.
MIMIC-III is provided by PhysioNet~\cite{PhysioBank}, and is a free database that collects health-related data from hospitals.
MIMIC-III contains data collected from two systems of Carevue and Metavision, which have different characteristics and are treated as heterogeneous datasets.

In the following experiments, we select SpO2, respiratory rate (RR), and mean arterial blood pressure (BP) as prediction targets (3 separate prediction tasks).
For the prediction features, we select heart rate, respiratory rate, arterial BP mean, NBP mean, and temperature in the Carevue dataset ($N = 5$), and select heart rate, respiratory rate, non-invasive BP mean, and arterial BP mean in the Metavision dataset ($N = 4$).
For real-life implementation, we only select indicators that could be measured by wearable devices as prediction features.

We filter out data of users obtaining less than 5 records of the prediction target, and the number of users in each dataset and prediction task is listed in Table~\ref{tab:NData}.
In addition, we divide the dataset into training (60\%), validation (20\%), and testing (20\%) based on the number of users, and the numbers of data instances are also listed in Table~\ref{tab:NData}.

\begin{table}[t]
\caption{Numbers of users and data in each dataset and prediction task}
\centering
\begin{tabular}{|c|c|c|c|c|}
\hline
\multicolumn{5}{|c|}{Carevue}\\
\hline
Target  & Users   & Training & Validation & Testing \\
\hline
SpO2    & 114,119 & 119,441 & 36,557 & 33,528\\
RR      & 107,084 & 119,305 & 36,546 & 32,814\\
BB      &  82,606 &  98,737 & 29,996 & 27,469\\
\hline
\multicolumn{5}{c}{}\\
\hline
\multicolumn{5}{|c|}{Metavision}\\
\hline
Target  & Users   & Training & Validation & Testing \\
\hline
SpO2    & 229,620 & 299,705 & 78,766 & 86,854\\
RR      & 233,646 & 304,312 & 78,458 & 86,070\\
BB      &  89,826 & 194,460 & 50,049 & 45,449\\
\hline
\end{tabular}
\label{tab:NData}
\end{table}

In this paper, 3 benchmark systems are adopted for comparison, including the multi-layer perceptron (MLP) and the state-of-the-art systems BIBE~\cite{SO-Pred}.
BIBE~\cite{SO-Pred} designed a convolutional neural network (CNN) based feature extractor with a semi-supervised pre-training mechanism to predict SpO2.
All benchmark systems take dense feature matrices as inputs.
For a fair comparison, we make all systems have a similar number of parameters in the neural network.
In the following experiments, the window size $W$ is set to 3, and each experiment is repeated 10 times to obtain the average performance.
All systems are trained for 100 epochs with L2 loss (MSE) and Adam optimizer (learning rate 0.01).
The save-best mechanism is adopted to save the model with the lowest MSE on the validation set.
The codes are written in python 3.8.12 with the PyTorch library, and are executed on Intel(R) Core(TM) i7-9700 CPU @ 3.00GHZ with 16GB RAM and an NVIDIA GeForce RTX 2080 Ti with 11GB GDDR6.

\begin{table}[t]
\caption{MSE of systems on the Carevue dataset}
\centering
\begin{tabular}{|c|c|c|c|c|}
\hline
Task  & MLP    &   BIBE & HTPS \\
\hline
SpO2  &  57.90 &  14.74 &  12.38\\
\hline
RR    &  47.65 &  44.55 &  44.26\\
\hline
BB    & 567.18 & 395.75 & 329.44\\
\hline
\end{tabular}
\label{tab:PEC}
\end{table}

\subsection{Prediction Evaluation}   \label{sec:Exp-PE}
Table~\ref{tab:PEC} lists the testing MSE of systems predicting SpO2, RR, and BP on the Carevue dataset.
Experimental results show that the proposed HTPS outperforms the benchmark systems among all tasks, and has the lowest prediction errors, especially in SpO2 and BP.
In SpO2 prediction, the proposed HTPS reduces MSE by 78.6\% and 16.0\% compared to MLP and BIBE.
In BP prediction, the proposed HTPS reduces MSE by 41.9\% and 30.2\% compared to MLP and BIBE.

Table~\ref{tab:PEM} illustrates the testing MSE of systems predicting SpO2, RR, and BP on the Metavision dataset.
Experimental results show that the benchmark systems perform the worst in SpO2 prediction, and obtain extremely large testing MSEs of 252.97 and 6257.39, respectively.
Contrarily, HTPS obtains a low and stable MSE of 29.72.
In RR prediction, the proposed HTPS reduces MSE by 57.0\% and 3.1\% compared to MLP and BIBE.
In BP prediction, the proposed HTPS reduces MSE by 3.1\% and 1.8\% compared to MLP and BIBE.

In summary, state-of-the-art BIBE systems perform well on almost all tasks and datasets, but sometimes perform poorly with extremely large testing MSEs.
On the other hand, HTPS achieves outstanding and stable performance under all prediction tasks and datasets, and can reduce the testing MSE at most by 78.6\% and 30.2\% compared to MLP and BIBE, respectively.
This phenomenon demonstrates that heterogeneous data has a negative impact on convergence, aligned with observation of~\cite{HData2}, while the proposed HTPS prevents the issue and achieves stable convergence and excellent performance in heterogeneous data.

\begin{table}[t]
\caption{MSE of systems on the Metavision dataset}
\centering
\begin{tabular}{|c|c|c|c|c|}
\hline
Task  & MLP     &    BIBE &    HTPS\\
\hline
SpO2  &  252.97 & 6257.39 &   29.72\\
\hline
RR    &   82.02 &   36.40 &   35.28\\
\hline
BB    & 3055.18 & 3011.89 & 2959.16\\
\hline
\end{tabular}
\label{tab:PEM}
\end{table}

\subsection{Ablation Studies}   \label{sec:Exp-AS}
In this section, we execute ablation studies to evaluate the effect of each module in the HTPS.
We compare four versions of HTPS, including MLP, AE, AEDS, and AEDST.
MLP is a baseline model that shares similar fully-connected layers with the prediction network of the HTPS.
DEN is an MLP system with the proposed dense embedding network, while DSEN is a DEN system with the sparse embedding network.
DSENT is the full version of HTPS, including dense embeddings, sparse embeddings, and prediction networks with heterogeneous transferring.

Experimental results demonstrate that as more networks and mechanisms are applied, the MSEs of all prediction tasks gradually decrease on the Metavision dataset, as shown in Fig.~\ref{fig:ABM}.
Experimental results show that the dense embedding network contributes the most to the improvement.
Transferring knowledge from the Carevue to the Metavision dataset can improve predictions.

\begin{figure}[t] 
  \centering
  \includegraphics[width=3.5in]{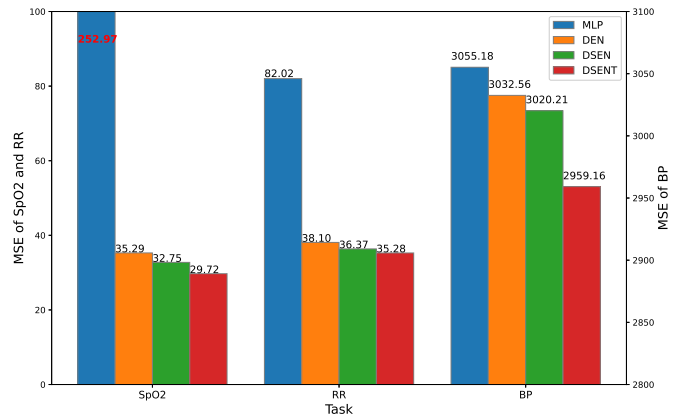}
  \caption{Ablation studies on the Metavision dataset}
  \label{fig:ABM}
\end{figure}

\section{Conclusion}   \label{sec:Con}
Medical Internet of Things (MIoT) leads to revolutionary improvements in medical services, involving the development of smart hospitals, IoT sensing, and diagnosis assisting, also known as smart healthcare.
With the big healthcare data, data mining and machine learning can assist wellness management and intelligent diagnosis, and achieve the P4-medicine.
However, healthcare data has high sparsity and heterogeneity.
How to make full use of heterogeneous healthcare data is still an unsolved problem.
In this paper, we propose a Heterogeneous Transferring Prediction System (HTPS).
Feature engineering in HTPS transforms the dataset into sparse and dense feature matrices, which are inputs of the designed sparse and dense embedding networks.
Autoencoders in the dense embedding networks not only embed the features but also transfer knowledge from heterogeneous models and datasets.
Experimental results show that the proposed HTPS outperforms the benchmark systems on various prediction tasks and datasets.
Ablation studies are executed to present the effectiveness of each designed mechanism in HTPS.
Experimental results demonstrate the negative impact of heterogeneous data on benchmark systems and the high transferability of the proposed HTPS.
In the future, we will employ federated learning to improve efficiency and will apply the proposed system in various research fields.

\section*{Acknowledgments}
This work is partially supported by the National Centre for Research and Development under the project Automated Guided Vehicles integrated with Collaborative Robots for Smart Industry Perspective and the Project Contract no. is: NOR/POLNOR/CoBotAGV/0027/2019-00.

\IEEEtriggeratref{7}
\bibliographystyle{IEEEtran}
\bibliography{refs}

\end{document}